\pdfoutput=1
\documentclass[11pt]{article}
\usepackage{overpic}
\usepackage{float}

\newif\ifcomment\commenttrue
\usepackage[a-1b]{pdfx}

\usepackage{framed}
\usepackage{lmodern}
\usepackage{mdwlist}
\usepackage{siunitx}
\usepackage{latexsym}
\usepackage{colortbl}
\usepackage{xcolor}
\usepackage{nicefrac}
\usepackage{booktabs}
\usepackage{fnpct}
\usepackage{amsfonts}
\usepackage[T1]{fontenc}
\usepackage{bold-extra}
\usepackage{amsmath}
\usepackage{amssymb}
\usepackage{bm}
\usepackage{graphicx}
\usepackage{mathtools}
\usepackage{microtype}
\usepackage{multirow}
\usepackage{multicol}
\usepackage{xpatch}
\usepackage{latexsym,comment}
\usepackage[normalem]{ulem}

\newcommand*{\missingreference}{{\Huge \colorbox{red}{?reference?}}}
\newcommand*{\missingcitation}{{\Huge \colorbox{red}{?citation?}}}

\makeatletter
\xpatchcmd{\@setref}{\bfseries}{\missingreference}{}{}
\def\@citex[#1]#2{\leavevmode
    \let\@citea\@empty
    \@cite{\@for\@citeb:=#2\do
        {\@citea\def\@citea{,\penalty\@m\ }%
            \edef\@citeb{\expandafter\@firstofone\@citeb\@empty}%
            \if@filesw\immediate\write\@auxout{\string\citation{\@citeb}}\fi
            \@ifundefined{b@\@citeb}{\hbox{\reset@font\missingcitation}%
                \G@refundefinedtrue
                \@latex@warning
                {Citation `\@citeb' on page \thepage \space undefined}}%
            {\@cite@ofmt{\csname b@\@citeb\endcsname}}}}{#1}}
\makeatother

\newcommand{\model}[0]{VibeJam\xspace}

\newcommand{\mm}[0]{\textsc{llm}\xspace}

\newcommand{\gem}[1]{\mbox{\textsc{gem}}}
\newcommand{\abr}[1]{\textsc{#1}\xspace}

\newcommand{\hidetext}[1]{}
\newcommand{\ignore}[1]{}

\ifcomment
    \newcommand{\pinaforecomment}[3]{\colorbox{#1}{\parbox{.8\linewidth}{#2: #3}}}

    \newcommand{\prtodo}[1]{\pinaforecomment{lightblue}{pr}{#1}}
    \newcommand{\prtodoi}[1]{\pinaforecomment{lightblue}{pr}{#1}}
\else
    \newcommand{\pinaforecomment}[3]{}
    \newcommand{\prtodo}[1]{}
    \newcommand{\prtodoi}[1]{}
\fi

\newcommand{\smallurl}[1]{ \begin{tiny}\url{#1}\end{tiny}}

\definecolor{lightblue}{HTML}{3cc7ea}
\definecolor{CUgold}{HTML}{CFB87C}
\definecolor{grey}{rgb}{0.95,0.95,0.95}
\definecolor{ceil}{rgb}{0.57, 0.63, 0.81}
\definecolor{UMDred}{HTML}{ed1c24}
\definecolor{UMDyellow}{HTML}{ffc20e}

\newcommand{\nlp}[0]{\abr{nlp}}
\newcommand{\ai}[0]{\abr{ai}}

\usepackage[]{acl2023}

\usepackage{xspace}
\usepackage{xcolor}
\usepackage[utf8]{inputenc}
\usepackage{pgfplots}
\usepackage{dsfont}
\DeclareUnicodeCharacter{2212}{−}
\usepgfplotslibrary{groupplots,dateplot}
\usetikzlibrary{patterns,shapes.arrows}
\pgfplotsset{compat=newest}

\definecolor{myblue}{RGB}{30,90,200}

\usepackage{tikzscale}
\usepackage{relsize}
\usepackage{amsmath,amssymb}
\newcommand{\probP}{\text{I\kern-0.15em P}}

\usepackage{multirow, colortbl}

\usepackage{tabularx,booktabs}
\usepackage{makecell}

\usepackage[normalem]{ulem}
\useunder{\uline}{\ul}{}

\usepackage{graphicx}

\definecolor{ablation6}{HTML}{fcefed}
\definecolor{ablation_tie}{HTML}{fce3e1}

\definecolor{ablation5}{HTML}{fcd8d4}
\definecolor{ablation4}{HTML}{FBC3BC}
\definecolor{ablation3}{HTML}{F7A399}
\definecolor{ablation2}{HTML}{F38375}
\definecolor{ablation1}{HTML}{EF6351}

\newcommand{\inlinecode}[1]{%
    \begin{tikzpicture}[baseline=0ex]%
         \node[anchor=base,%
         text height=0.7em,%
         text depth=0.7ex,%
         inner ysep=0pt,%
         draw=lightgray!50,%
         fill=lightgray!50,%
         rounded corners=2pt] at (0,0) {\footnotesize\texttt{#1}};%
    \end{tikzpicture}%
}

\definecolor{OliveGreen}{rgb}{0.05, 0.75, 0.24}
\definecolor{BrickRed}{rgb}{0.8, 0.25, 0.33}

\usepackage[]{algpseudocode}
\usepackage[]{algorithm}
\usepackage{float}
\algtext*{EndFor}%
\algtext*{EndProcedure}%

\usepackage{booktabs}
\usepackage[normalem]{ulem}
\useunder{\uline}{\ul}{}

\usepackage{times}
\usepackage{latexsym}
\usepackage{adjustbox}

\usepackage[T1]{fontenc}
\usepackage{amsmath}
\usepackage{amssymb}
\usepackage{booktabs}
\usepackage{tikzscale}
\usepackage{amsmath}

\usepackage{multirow, colortbl}

\usepackage{tabularx,booktabs}
\usepackage{makecell}
\usepackage{multirow}
\usepackage{scalerel,xparse}
\usepackage{cleveref}
\usepackage{microtype}
\usepackage[most]{tcolorbox}

\usepackage{enumitem}
\crefformat{section}{\S#2#1#3}
\crefformat{subsection}{\S#2#1#3}
\crefformat{subsubsection}{\S#2#1#3}

\definecolor{bggray}{rgb}{0.95, 0.95, 0.95}
\usepackage[%
    framemethod=tikz,
    skipbelow=\topskip,
    skipabove=\topskip
]{mdframed}
\mdfsetup{%
    leftmargin=0pt,
    rightmargin=0pt,
    backgroundcolor=bggray,
    middlelinecolor=black,
    roundcorner=3
}

\definecolor{SkyBlue}{rgb}{0.53, 0.81, 0.92}

\newtcolorbox[
  list inside=prompt,
  auto counter,
  number within=section
]{prompt}[1][]{%
  enhanced,
  float*=t,                 % ← span both columns (top). Use float*=htbp if you want more placement options
  colbacktitle=black!60,
  fonttitle=\small,
  coltitle=white,
  fontupper=\footnotesize,
  boxsep=4pt,
  left=0pt, right=0pt, top=0pt, bottom=0pt,
  boxrule=1pt,
  width=\textwidth,          % full page width inside the spanning float
  enlarge left by=0mm,
  enlarge right by=0mm,
  listing only,
  listing options={
    basicstyle=\ttfamily\footnotesize,
    breaklines=true,
    breakatwhitespace=true,
    language=json
  },
  #1,
}

\definecolor{circlered}{HTML}{fd6e51}
\definecolor{circleblue}{HTML}{2596be}
\definecolor{circleorange}{HTML}{f6b621}

\newcommand{\circled}[2][circlered]{%
  \tikz[baseline=(char.base)]{
    \node[
      shape=circle,
      draw,
      fill=#1,
      inner sep=1pt
    ] (char) {#2};}}

\newcommand{\circledorange}[2][circleorange]{%
  \tikz[baseline=(char.base)]{
    \node[
      shape=circle,
      draw,
      fill=#1,
      inner sep=1pt
    ] (char) {#2};}}

\newcommand{\circledblue}[2][circleblue]{%
  \tikz[baseline=(char.base)]{
    \node[
      shape=circle,
      draw,
      fill=#1,
      inner sep=1pt
    ] (char) {#2};}}

\newtcolorbox[
  list inside=trace,
  auto counter,
  number within=section
]{trace}[1][]{%
  enhanced,
  float*=t,
  colback=blue!5,             % light blue background
  colbacktitle=blue!60!black, % darker blue title bar
  colframe=blue!60!black,     % matching border
  fonttitle=\small,
  coltitle=white,
  fontupper=\footnotesize,
  boxsep=4pt,
  left=0pt, right=0pt, top=0pt, bottom=0pt,
  boxrule=1pt,
  width=\textwidth,
  enlarge left by=0mm,
  enlarge right by=0mm,
  listing only,
  listing options={
    basicstyle=\ttfamily\footnotesize,
    breaklines=true,
    breakatwhitespace=true,
    language=json
  },
  #1,
}

\definecolor{UMDred}{HTML}{ed1c24}

\usepackage{color-edits}
\addauthor{vc}{blue}

\definecolor{yellowcolor}{HTML}{ffc20e}
\definecolor{redcolor}{HTML}{e99999}
\definecolor{orangecolor}{HTML}{f6b26b}
\definecolor{yellowcolor}{HTML}{ffd966}
\definecolor{bluecolor}{HTML}{a0c5e8}
\definecolor{purplecolor}{HTML}{d9d2e9}

\usepackage{tikz}

\usepackage{xcolor}
\definecolor{highlight}{HTML}{FFAE02}

\title{
\raisebox{-0.3333em}{\includegraphics[height=1.5em]{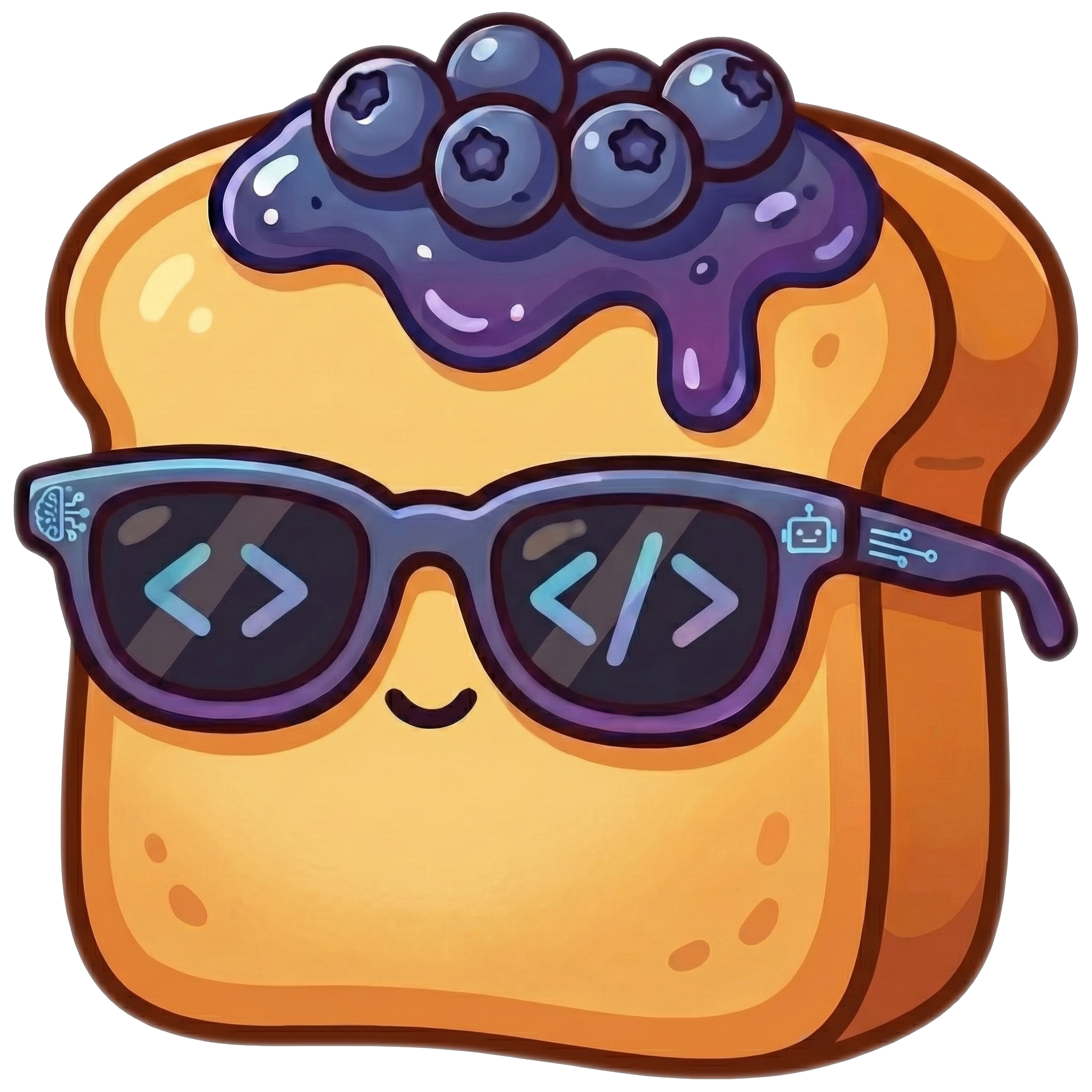}} 
\model{}:
A User Study Platform for Web Development with Agents
}

\renewcommand{\thefootnote}{*}

\newcommand{\authorSpacing}{0.8cm}
\author{
\textbf{Nishant Balepur}$^{1, 2}$\hspace{\authorSpacing}
\textbf{Connor Baumler}$^{1}$ \hspace{\authorSpacing}
\textbf{Valerie Chen}$^{3}$ \hspace{\authorSpacing}\\
\textbf{Eunsol Choi}$^{2}$ \hspace{\authorSpacing}
\textbf{Rachel Rudinger}$^{1}$ \hspace{\authorSpacing}
\textbf{Jordan Boyd-Graber}$^{4}$\footnotemark  \\[0.5em]
$^{1}$University of Maryland \hspace{0.3cm}
$^{2}$New York University \hspace{0.3cm}\\[0.25em]
$^{3}$Carnegie Mellon University \hspace{0.3cm}
$^{4}$Nanyang Technological University
\\[0.25em]
\raisebox{-0.3333em}{\includegraphics[height=1.5em]{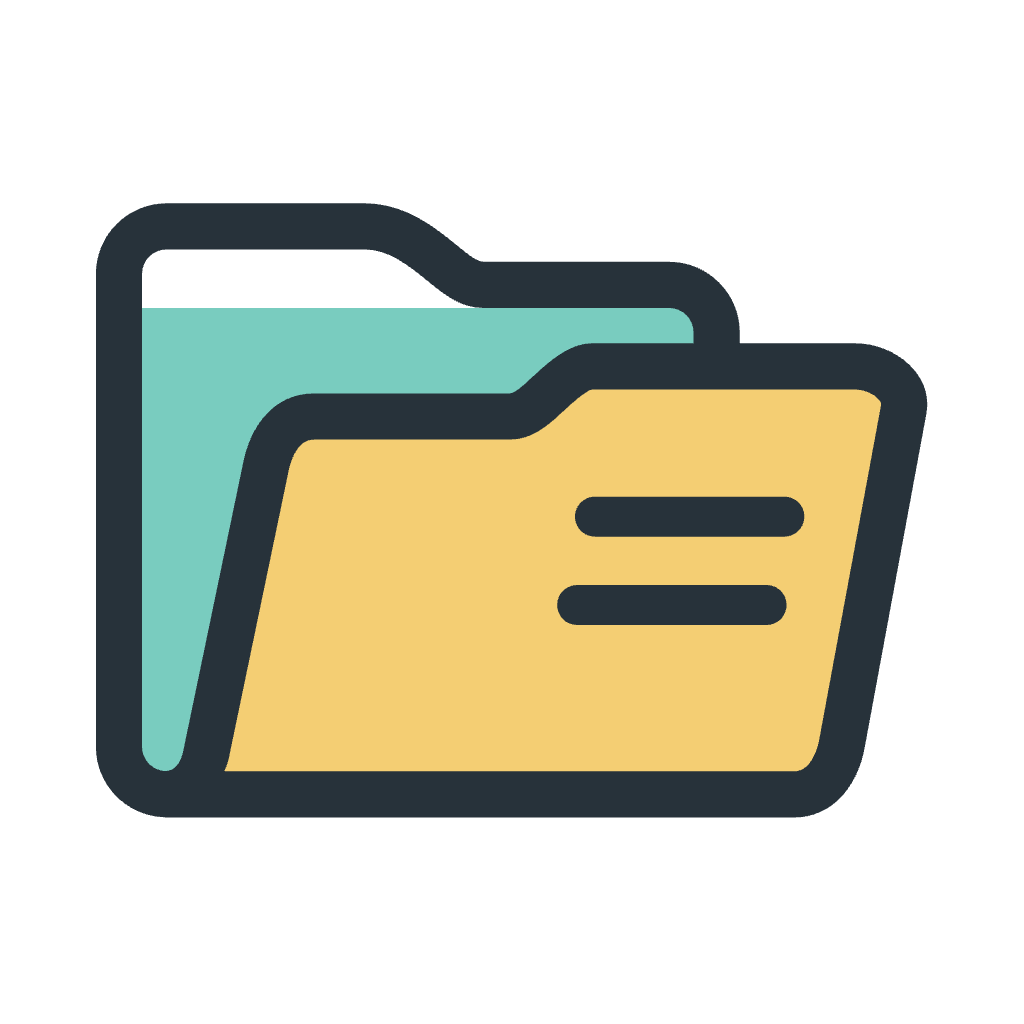}} \textbf{Code:} \url{https://github.com/nbalepur/vibe-jam}\\[0em]
\raisebox{-0.3333em}{\includegraphics[height=1.5em]{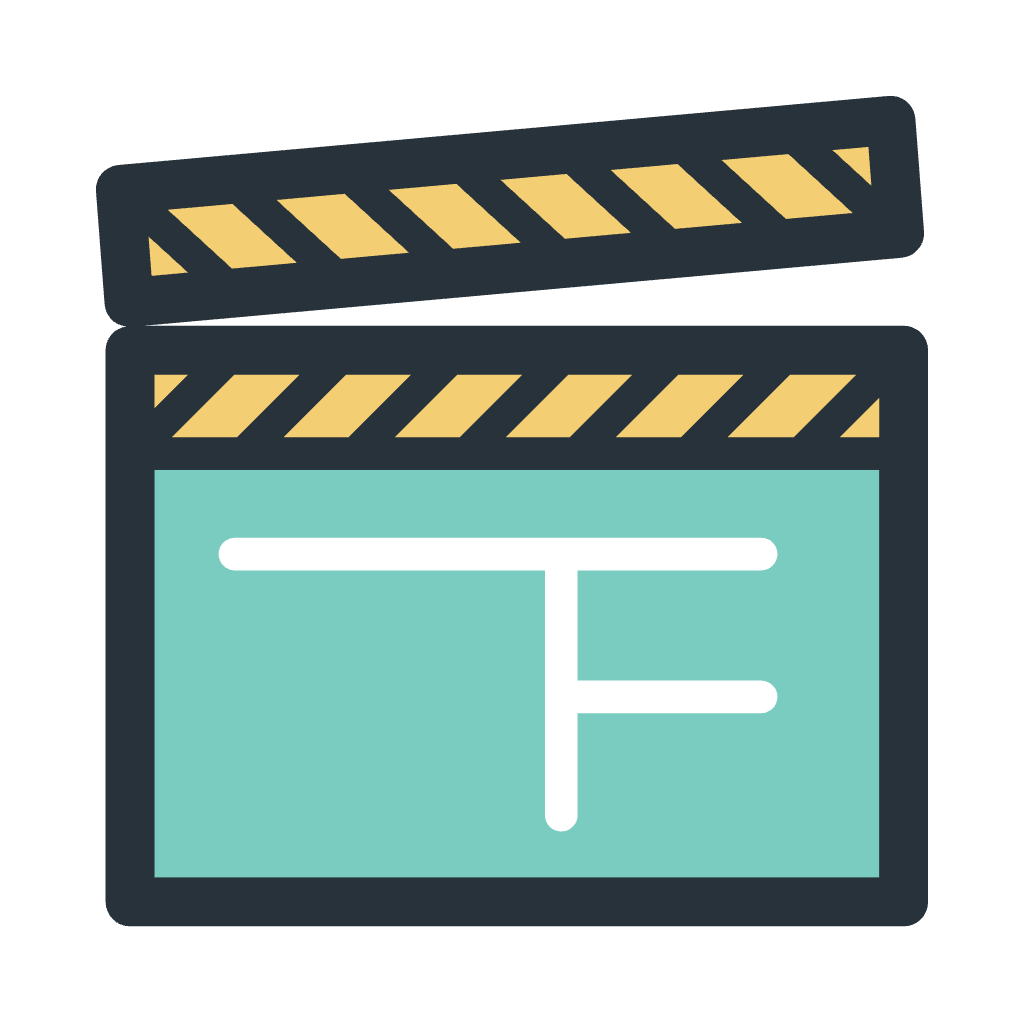}} \textbf{Video Demo:} \url{https://youtu.be/Kg9kqKWszw8}\\[0.5em]
\texttt{nbalepur@umd.edu} \hspace{0.5em} \texttt{jordan.ying@ntu.edu.sg}
}

\begin{document}
\maketitle

\begin{abstract}
Programming with \ai{} is increasingly~agentic---users prompt \mm{}s to directly edit their code and review the changes---with adoption growing especially for web development tasks.
Despite this growth, most \nlp{} work uses offline evaluation and lacks support for online studies, losing insights into how programmers truly use coding agents.
We release \model{}, a browser-based user study platform for users to collaborate with \ai{} agents to develop websites.
\model{} enables agent customization and uses the open-source Aider agent by default, and to mirror downstream use, we add diff review, chat and plan modes, and live website previews.
In a pilot study with $55$ released, game-based website creation tasks, five experienced \ai{} programmers rate our system as fun, simple, and resembling commercial tools, while 13 junior students use \model{} to make websites of higher quality~than agents in the same task.
We open-source \model{} to spur extensions~and~support studies on how coding agents can help users.
\footnotetext{Work done at University of Maryland}
% \vccomment{overall, motivation of work in abstract could be worded a little more strongly / clearly. i think we talk a lot about the features but its not clear why i need them}
\end{abstract}

\renewcommand{\thefootnote}{\arabic{footnote}}

\section{Introduction: Agentic Coding's Boom} \label{section:intro}

Advances in Large Language Models (\mm{}s) have shifted \ai-assisted coding from autocomplete suggestions to agentic workflows \cite{yang-etal-2025-code}, where a user prompts agents to directly edit code and reviews edits. 
The growth of tools~like Cursor mark~this shift, where even novices use~agents~to write working code~\cite[i.e., vibe code]{hu2026human}.

\setlength{\fboxrule}{1.5pt}

\begin{figure*}[t!]
    \centering
    \includegraphics[width=\linewidth]{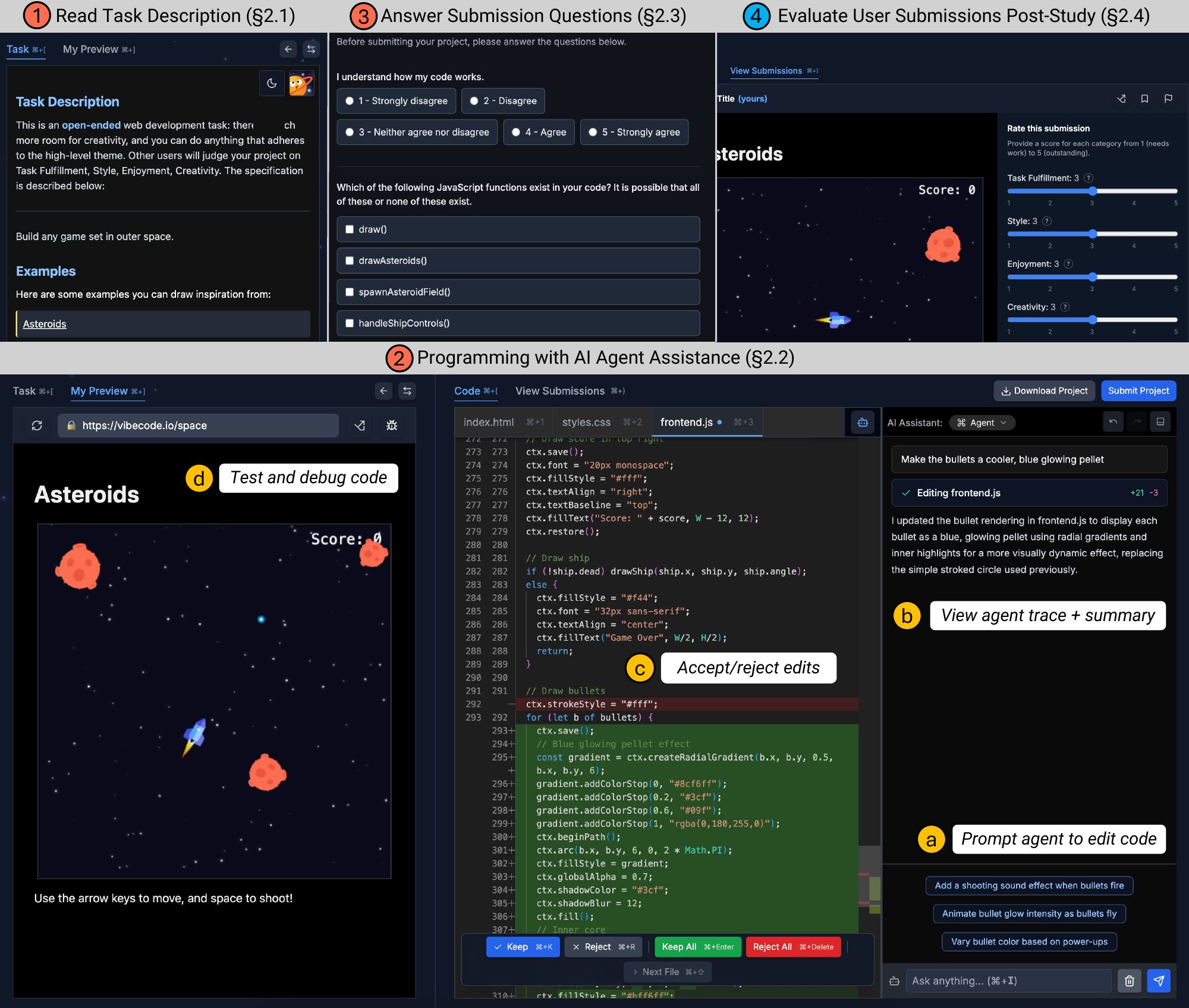}
    \vspace{-4ex}
    \caption{User study workflow in \model{} for the task: ``Make a game in outer space''. Users read the description of a website or function task (\circled{1}) to complete in a VSCode-style editor with an \ai{} agent (\circled{2}), along with chat and plan modes (Figures~\ref{fig:chat}, \ref{fig:plan}).
    Users prompt the agent to modify files (\circledorange{a}) and see its execution trace (\circledorange{b}), which can be reviewed via a diff-based editor (\circledorange{c}).
    Users can test their code with a live website preview (\circledorange{d}).
    After submitting code, users answer fixed questions or questions generated based on their code (\circled{3}).
    Post-study, researchers can mark specific users as annotators to rate all submissions (\circledblue{4}). We evaluate \model{} through two user studies in \cref{section:results}.
    }
    % \vspace{-1ex}
    \label{fig:vibejam}
\end{figure*}

\nlp progress on coding agents hinges on offline benchmarks \cite{chen2021codex}, but to ensure these scores translate to online benefits in deployment, we must assess them in ways mirroring how users work with them \cite{balepur-etal-2025-best}.
A key~reason online studies are limited is a lack of system support; existing coding interfaces are either not public \cite{shen2026ai}, require user setup \cite[e.g., VSCode plugins]{yan2024ivie}, or only support older \ai features \cite[e.g., autocomplete for task completion]{mozannar2024realhumaneval}, precluding user studies and online evaluations of agents.
%\vccomment{last part of sentence implies that vibejam supports testing any coding agent}

We release \textbf{\model{}}\footnote{We use ``Jam'' as a reference to ``Game Jam'', reflecting our hope to inspire fun, creative submissions like games (\cref{subsection:pilot}).} (Figure~\ref{fig:vibejam}), a customizable platform for agentic coding user studies that requires no participant setup.
Informed by coding reports \citep{anthropic2025impact} and a survey with eight \ai{} coding researchers (\cref{section:formative}), we initially build \model{} for web development with features researchers value (e.g., multiple \ai{} modes, model customization), but open-source it to spur novel extensions.

Instead of unrealistically aiming to beat commercial coding agents, we design \model{} to mirror such agents for user studies with downstream realism.
In \model{}'s typical study workflow,~users design websites (\circled{1}, \cref{subsection:tasks}) by collaborating with an \ai{} that can switch between agent, chatbot, and plan modes (\circled{2}, \cref{subsection:ai}).
The \ai{} is configurable but uses the open-sourced \textsc{Aider} framework \cite{aider} by default for agents and a prompted \mm{} for chat and plan modes.
Users can prompt the agent (\circledorange{a}), read execution traces (\circledorange{b}), review the changes (\circledorange{c}), and debug with website previews (\circledorange{d}).
After every task,~users answer~questions (\circled{3}, \cref{subsection:submission}) that researchers can fix across users or tailor to a user's code via \mm{}s.
Researchers can also mark users as annotators to let them rate all user task submissions on custom metrics (\circledblue{4}, \cref{subsection:evaluation})---so \model{} helps researchers evaluate user code from their studies. 

% Deploying this workflow adds challenges in long-context management and safe execution of arbitrary code, which we address by integrating patch-based edits and context summarization, and by sandboxing all code in \inlinecode{iframes} or third-party~\textsc{api}s.
% \vccomment{a lot of the content feels way too detailed for intro, recommend checking other demo papers to see if they do the same}

% \connor{Maybe add number to fig 1 showing where in the interface the buttons for these things are and number this list to match?}

% \model{} includes features tailored for \nlp{} studies. \vccomment{need for these features are not super clear to me, second paragarph should motivate better}
% When users complete a task, we send their~code to an \textsc{api} endpoint to return post-submission questions, letting researchers specify fixed survey questions or enabling the generation of questions tailored to user code (\cref{subsection:submission}).
% We also add an evaluation pane for specially-marked annotator users to score submitted code (\cref{subsection:evaluation}).
% Finally, given \nlp{}'s lack~of data for easy-to-understand website tasks \cite{si-etal-2025-design2code}, we design and release 55 tasks on game development that can be loaded into \model{} (\cref{subsection:research}). \vccomment{why these tasks?}

% \vccomment{summary sentence should say smth like to validate vibejam, we had two goals..}

To evaluate \model{}'s value as a user study platform, we first run a pilot study where five users~create a game-based website in \model{}; they deem our system fun, easy to use, and mirroring commercial tools (\cref{subsection:pilot}).
We then run a more realistic study where 13 junior programmers make 22 game-based websites in \model{}, which end up~besting the~quality of agents alone (\cref{subsection:wild}).
These successful user study outcomes reveal that \model{} can support user-centered evaluations of~coding agents.

We release \model{} at a time where researchers want to study agents with users in the loop \cite{wang2025humansmissing}.
Toward this, we end with ways \model{} can support \nlp work in active areas like long-horizon execution, code quality, and \ai{} safety (\cref{section:conclusion}).

\section{A Formative Study to Ground \model{}} \label{section:formative}

We run a formative study \citep{Nielsen1993UsabilityE} to ensure we design \model{} to align with what researchers value.
Eight \nlp{}+coding researchers from distinct universities review an initial version of our demo, then take a survey to rate the importance of different file structure, \ai{} interaction, and configurability design choices.
We distill their feedback below.

\begin{figure}[t!]
    \centering
    \includegraphics[width=\linewidth]{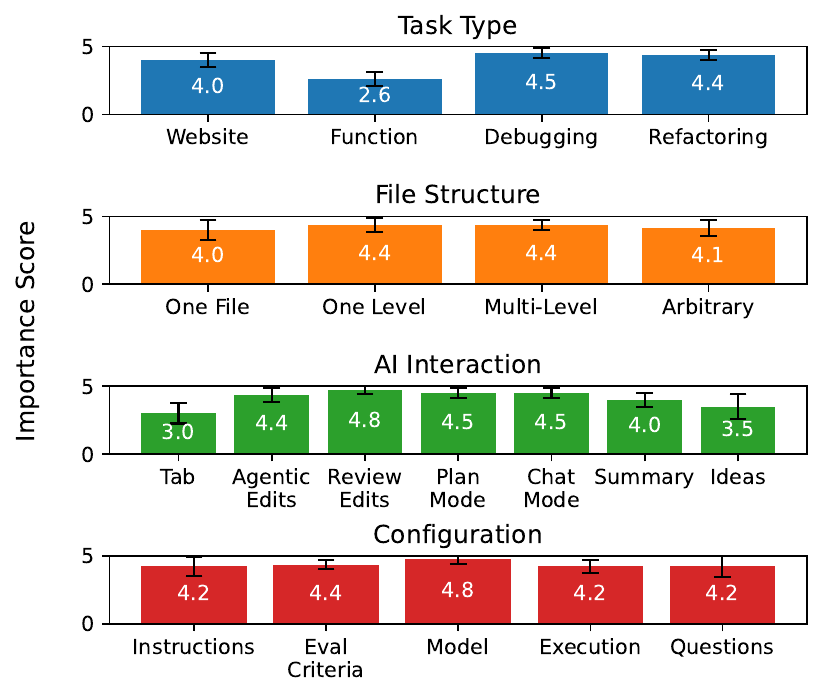}
    \vspace{-3ex}
    \caption{Results from our formative study: 1--5 importance scores from \ai{} coding researchers on features they find valuable for a user study platform for agentic coding, which we distill to inform \model{}'s design.}
    \label{fig:author}
    % \vspace{-1.5ex}
\end{figure}

Researchers rate website, debugging, and refactoring tasks as highly important but not function completion (Figure~\ref{fig:author}, blue), even though current \textsc{ui}s only support the latter \cite{mozannar2024realhumaneval}---leaving a gap \model{} is designed to address.
Each task type needs distinct affordances, so we initially scope \model{} to a single task rather than aiming to support all coding workflows.
We choose web development for two reasons.
First, \citet{anthropic2025impact} show web development languages (e.g., JavaScript, HTML) are the most common use case for Claude's coding agents---encompassing over $60\%$ of user requests.
Second, \citet{robbes2026agentic} find high adoption of coding agents for website tasks on GitHub.

Researchers found one level of files sufficient,~so we support that scope in \cref{subsection:tasks} to reduce the complexity of our interface.
They rated~agentic edits as more important than tab autocomplete (Figure~\ref{fig:author}, green), confirming the need for an agentic coding system beyond existing \textsc{ui}s (\cref{section:related_work}).
They also highlighted the value of chat and plan modes, review mechanisms, and summaries, which we add~in \cref{subsection:ai}.
Lastly, all researchers desired customization (Figure~\ref{fig:author}, red), notably the ability to select the \mm{} backing coding agents and define custom evaluation metrics.
This inspired us to specifically target the latter via an evaluation pane where designated annotators can rate user-submitted websites (\cref{subsection:evaluation}).

Overall feedback was positive, and no one identified an existing system with similar features.
One researcher noted the potential of our platform for arena-style rankings~\citep{zheng2024judging}.
Another was ``\textit{very excited}'' to use \model{} and felt ``\textit{Better vibecode-studying tools would be SUPER useful!}''.
We consider this a positive sign to pursue \model{} and use this feedback to guide our implementation.

\section{\model{}: An Agentic Coding System} \label{section:method}

Informed by the \nlp{}+coding community, we now outline \model{}'s features (Figure~\ref{fig:vibejam}).
We describe tasks (\cref{subsection:tasks}), \ai{} tools (\cref{subsection:ai}), and debugging (\cref{subsection:testing}) in \model{}, followed by protocols for submitting code (\cref{subsection:submission}), scoring submissions (\cref{subsection:evaluation}), and features to support \textsc{ai} web development studies (\cref{subsection:research}).

\subsection{Task Types} \label{subsection:tasks}

Each task in \model{} has users produce a website via an input description, enabling tasks like recreating a site to pass test cases \citep{zhu2025frontendbench} or creatively designing a site under high-level themes \citep{fukumura2025can}.
Researchers can add a task as a database row with a task ID, files the user initially sees (e.g., \inlinecode{frontend.js}, \inlinecode{index.html}, \inlinecode{styles.css}), and a task description.
Upon picking a task, users can click the ``Task'' tab to view instructions on what to complete (Figure~\ref{fig:vibejam}, \circled{1}).

% \model{} is the first website UI where~agents edit users' code, so we currently only support predefined files for simplicity---and not custom files or multi-level repositories \cite{jimenez2024swebench}---but we discuss how \model{} could support this in \cref{subsection:author}.

% \model{} has two task types: \textit{1) Web Design:}~users make a website based on a description; and \textit{2) Function Completion:} users complete a Python function from a description to pass test cases.
% Both are popular use cases of \textsc{ai}-assisted coding tools \cite{anthropic2025impact} and are explored in \nlp{} benchmarks \cite{chen2021codex, si-etal-2025-design2code}, making them valuable for a user study platform like~\model{}. The task types also differ in subjectivity---web design is subjective \cite{moshagen2010facets}---enabling evaluation of diverse metrics like creativity via website ratings \cite{10.1145/3635636.3656201} and productivity via test case accuracy \cite{becker2025measuring}.

% \input{figures/task}

%All task types have the same \ai{} features (\cref{subsection:ai}) but differ in debugging tools (\cref{subsection:testing}), described next.

\subsection{Human-\textsc{ai} Interactions} \label{subsection:ai}

We center \model{} on the new paradigm of \textbf{agentic coding}---a user writes a prompt, an \ai{} agent modifies the codebase, and the user reviews the changes \cite{shen2025completion}---which has grown popular in Cursor, Claude Code, and Copilot \cite{robbes2026agentic}.
Even users with no coding experience are using said tools via vibe coding \cite{chou2025building}: prompting \ai{} to produce code with minimal review.
Given the expected productivity gains and potential risks of this shift \cite{anthropic2026agentic}, we ground \model{} on this interaction \cite{gomez2025human}.

We now describe \model{}'s agentic coding, split into writing (\cref{subsubsection:ai_execution}) and reviewing (\cref{subsubsection:human_review}) code.

\subsubsection {Writing Code with \ai{} Agents} \label{subsubsection:ai_execution}

Users program in \model{} via Monaco Editor~\cite{monaco-editor} from VSCode.
Users can write code manually, but informed by \cref{section:formative}, we add an \ai{} assistant with three modes: 1) \textit{agent}: directly edit users' code \citep{robbes2026agentic}; 2) \textit{chat}: answer questions about users' code \citep{mailach2025ok}; and 3) \textit{plan}: create a specification for complex requests, then let the agent in (1) execute the plan \citep{liu2026dive}.
Each \ai{} is configurable in an \textsc{api} endpoint so researchers can compare designs \cite{10.1145/3708359.3712104}.
We describe the three \ai{} modes next.

\vspace{0.33333em}

\begin{figure*}[t!]
    \centering
    \includegraphics[width=\linewidth]{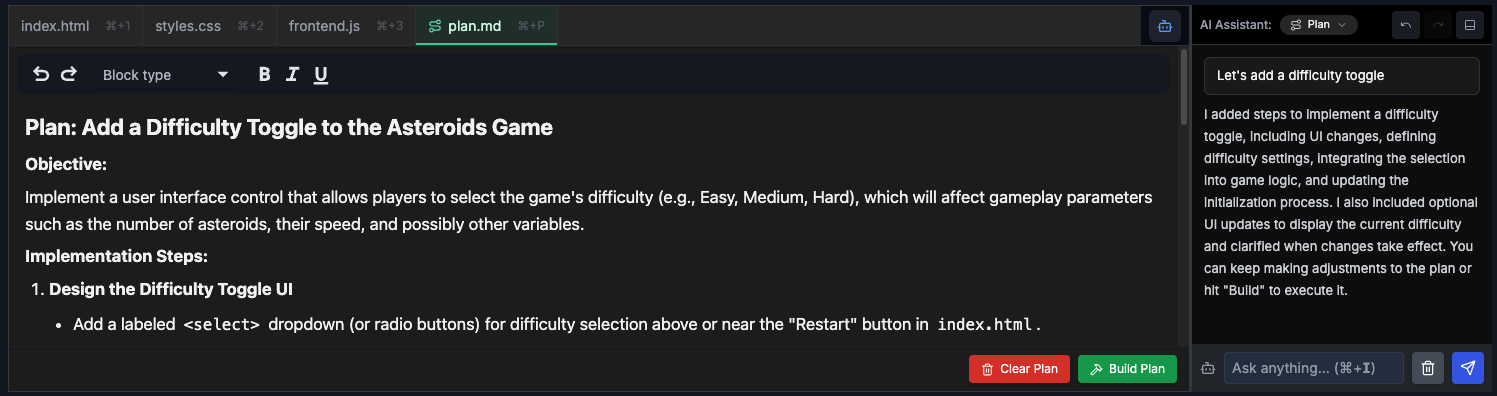}
    \vspace{-2ex}
    \caption{In Plan Mode, users write and adjust instructions for complex prompts, later given to the agent to execute.}
    % \vspace{-1.5ex}
    \label{fig:plan}
\end{figure*}

\noindent \textbf{Agent:} During agent mode, after the user submits a prompt (Figure~\ref{fig:vibejam}, \circledorange{a}), the agent writes the user's files from Monaco to a temporary directory on the backend, edits the files via the user's request, and returns the new files back to the Monaco editor.
We stream all execution edits: which files are being edited and how many lines were changed (Figure~\ref{fig:vibejam}, \circledorange{b}).
When edits end, we prompt an \mm{} to summarize changes and give follow-up ideas for ways users can improve their code (Prompt~\ref{prompt:summary_and_idea}).
Summaries are often used in agentic coding tools for comprehension \cite{wyrich202340}, while we expect follow-up idea generation to facilitate the development of long-horizon coding agents (\cref{section:conclusion}).

We implement the \textsc{Aider} \cite{aider} agent by default: an open-source library with 40k stars that supports popular \mm{} providers (OpenAI, Claude, local \mm{}s) and over 100 programming languages.
To manage errors stemming from long context windows \cite{sinha2025the}, we configure \textsc{Aider} to edit users' code by generating targeted diff patches versus regenerating files \cite{10.1109/ICSE48619.2023.00129} and summarize prior conversation history when it exceeds the model's context length \cite{chang2024booookscore}.
We also add a button that clears the conversation history so users can directly manage context length.

\vspace{0.33333em}

\begin{figure}[t!]
    \centering
    \includegraphics[width=\linewidth]{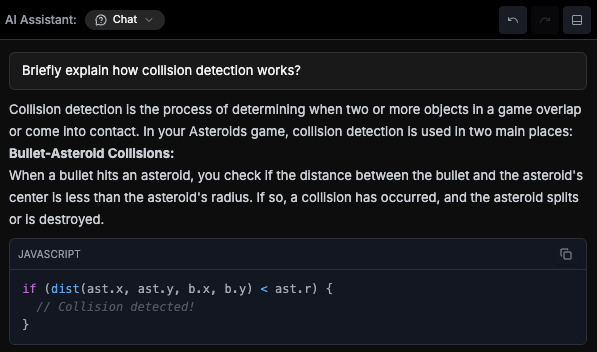}
    \vspace{-2ex}
    \caption{In Chat Mode, users ask questions about their code and view explanations with relevant code snippets.}
    % \vspace{-1.5ex}
    \label{fig:chat}
\end{figure}

\noindent \textbf{Chat:} Chat mode helps users who want to understand their code, such as during onboarding \citep{chen2025code}.
We add it via a prompt (Prompt~\ref{prompt:chat}) where an \mm{} reads the users' files, but only~returns text or code snippets in the assistant pane~without direct edits (Figure~\ref{fig:chat}). 
We render all produced code snippets with syntax highlighting for readability.

\vspace{0.33333em}

\noindent \textbf{Plan:} Plan mode is a new interaction mechanism where users can iterate with \ai{} to generate a long specification for a complex request (e.g., refactoring, new feature), then instruct an agent to execute said specification \cite{liu2026dive}.
We implement plan mode with an \mm{} that is prompted to generate a new plan or revise an existing plan based on the users' request (Prompt~\ref{prompt:plan}), then summarizes the changes it makes (Prompt~\ref{prompt:plan_summary}).
The plan renders in markdown and can be manually edited.
Users can then click ``Build`` on the plan to feed it to the \textsc{Aider} agent or clear it to restart (Figure~\ref{fig:plan}).

% When the user submits a prompt (Figure~\ref{fig:vibejam}, \circledorange{a}), \textsc{Aider} writes the user's files in Monaco to a temporary directory on the backend, edits the files via the user's request, and returns the new files back to Monaco.
% We stream all execution edits: which files are being edited and how many lines were changed (Figure~\ref{fig:vibejam}, \circledorange{b}).
% When edits end, we prompt an \mm{} to summarize changes and give follow-up ideas for ways users can improve their code (Prompt~\ref{prompt:summary_and_idea}).
% Summaries are often used in agentic coding tools for comprehension \cite{wyrich202340}, while we expect follow-up idea generation to facilitate the development of long-horizon coding agents (\cref{section:conclusion}).

\vspace{0.25em}

\subsubsection{Reviewing \textsc{ai}-Generated Code} \label{subsubsection:human_review}

In agent mode, users can review model-generated outputs in the form of ``Git diffs'', a desired feature in \cref{section:formative}.
Users see a dot on all files with \ai edits; upon navigating to one, they can view which lines were added/removed, and buttons to accept or reject the changes at the file or codebase level (Figure~\ref{fig:vibejam}, \circledorange{c}).

These changes are rendered when users preview their websites or run test cases (described in \cref{subsection:testing}), informing whether or not to accept them.
Users can also prompt the \ai{} to make new changes or manually edit their code before reviewing all changes.

\subsection{Testing and Debugging Code} \label{subsection:testing}

Debugging is a key part of software development \cite{zeller2009programs}, so we add support to help users identify and resolve issues.
We~include~a~``Preview'' tab that allows users to view their current site (Figure~\ref{fig:vibejam}, \circledorange{d}) and a console that relays all \inlinecode{console.log()} statements (Figure~\ref{fig:vibejam}, bottom left)---equivalent to \texttt{Inspect Element} in Chrome DevTools \cite{tanner2019poirot}.
To mitigate risks of running arbitrary user code, we sandbox all website code in an \inlinecode{iframe} without access to the parent interface.
% For function tasks, we include a ``Test Cases'' tab (Figure~\ref{fig:debug}) where users can test inputs or run public test cases.
% Consoles show all printed outputs, errors in users' code, and line numbers/tracebacks to locate issues.

% and execute all backend code via OneCompilerAPI\footnote{https://onecompiler.com/apis/pricing} which is fairly cheap (1000 reqs/month for free and 10,000 reqs/month for \$5).
% We also add local Python execution via \inlinecode{exec()} for development without \textsc{api} costs.

% \input{figures/submit}

\subsection{Code Submission} \label{subsection:submission}

When users submit their code to complete a task, we call a backend \textsc{api} that receives the user's code and returns questions to ask the user (Figure~\ref{fig:vibejam}, \circled{3}).
\model{} supports~multiple-choice, multiple-select, and free-response questions.
This lets researchers write fixed questions for all users post-submission \cite{hart1988development}, while adding the potential to tailor questions per user via question generation tools \cite{10.1017/S1351324906004177}.
We find the latter promising for constructs like comprehension that require writing questions per-user \cite{wyrich202340}, which we explore in Appendix~\ref{subsection:question}.
To incentivize users working toward outputs they may truly use, we add a download button to save a copy of the website.
The files also come with a script to publicly deploy the code for free on GitHub Pages.

\subsection{Submission Evaluation} \label{subsection:evaluation}

Along with supporting agentic coding workflows, \model{} adds support for evaluating user code.
Researchers can designate certain users as annotators, allowing them to view all task submissions.
Annotators can~open submissions and access the website preview in \cref{subsection:testing} to test users' code (Figure~\ref{fig:vibejam}, \circledblue{4}).
% : a preview for websites and a test case panel for function tasks.
% The latter loads public along with private test~cases.

Researchers can define custom evaluation metrics based on study goals, like creativity and code quality for website tasks.
This feature is meant for annotation, but as discussed in the formative (\cref{section:formative}), it could enable gamification if users also annotate via in-the-wild peer voting \cite{indriasari2023gamification}.

\subsection{Supporting Research Studies} \label{subsection:research}

We ensure \model{} is easy to use.
We simplify \textsc{ui} setup via bash scripts that: 1) install dependencies; 2) launch the system; and 3) load \textsc{json} task data.

A login system links all activity to user accounts, enabling cross-session studies.
We log most interaction data: \ai{} prompts/outputs, accept decisions on \textsc{ai} code, user code edits, submission question~responses, and annotator ratings.
To onboard users new to \ai{} coding, we release tutorial instructions and videos.
To facilitate studies, we add templates for consent forms, compensation, and contact.

Finally, there are few website tasks to support user studies on web development \cite{si-etal-2025-design2code}, so we curate 55 for future work.
We~desire~tasks that are easy-to-understand and fun for users, so we focus on game development \citep{chi2026gamedevbenchevaluatingagenticcapabilities}---hence our system's name of ``VibeJam'' (i.e., Vibe Coding plus Game Jams, a type of hack-a-thon for game development).
Twelve \nlp{} students ideate 23 replication tasks (e.g., re-create ``tic-tac-toe'') and 22 open-ended tasks (e.g., design a game in outer space).
One author reviewed all tasks and removed infeasible ones (e.g., tasks calling external \textsc{api}s).

% \connor{maybe doesn't matter for this venue, but is the 55 list ripped directly from the google sheet or did you cut some out, merge or alter them, etc?}

\section{\model{} in Practice: A Test Case Suite} \label{section:results}

To test \model{}'s practical value, we run two user studies.
First, five experienced \ai{} programmers~design websites in \model{}, who note our system's simplicity, efficacy, and similarity to commercial tools (\cref{subsection:pilot}).
Next, $13$ students use \model{} to~build $22$ websites better than agents alone---motivating \model{}'s value for human-centered evaluations.

\subsection{Pilot: \model{} Mirrors Commercial Tools} \label{subsection:pilot}

To validate \model{} mirrors how programmers use coding agents, we have five CS students with past \ai coding experience each make a website in \model{} with \textsc{Aider} (GPT-4.1), completing three of our tasks in \cref{subsection:research}: recreating tic-tac-toe, snake, and connect four.
After, they complete a survey with seven 1--5 Likert agreement questions across three dimensions: \textbf{1) Website Perceptions} (perceived quality and speed compared to coding without \ai); \textbf{2) AI Interactions} (ease of prompting, reviewing changes, and previewing the site); and \textbf{3) Experience} (fun and similarity to existing tools).
Users worked on websites until they were personally satisfied with the output, which took 15--60 minutes.
% \connor{not sure if reviewers will be annoying about asteroids not being one of the three pilot tasks. you should also make it clear whether this website came from the pilot or if it's one that you made yourself}

\begin{figure}[t!]
    \centering
    \includegraphics[width=\linewidth]{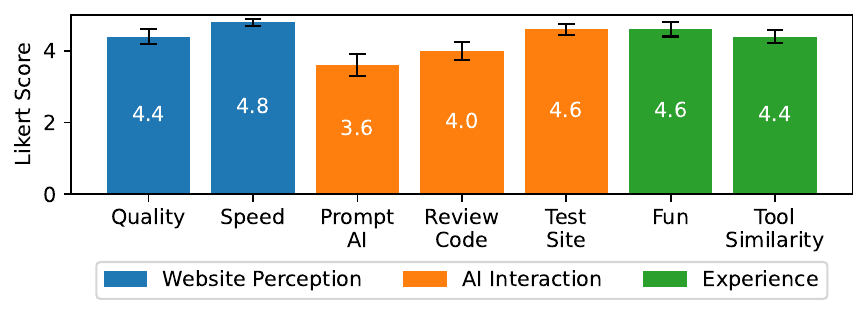}
    \vspace{-3ex}
    \caption{User feedback from our pilot survey. Users felt they built higher-quality websites in \model{} more quickly than without \ai, and stated that our system was easy to use, engaging, and similar to commercial tools.}
    \label{fig:user}
    % \vspace{-1.5ex}
\end{figure}

Users said \model{} helped them design higher-quality websites more quickly than they would have without \ai{} (Figure~\ref{fig:user}).
Reviewing~code~and testing websites was easy, but the agent did~not~follow all prompts perfectly;
our analyses found~\model{}---like many coding agents---struggled on complex, multi-step prompts, aligning with research showing the necessity of decomposition for success in agentic coding \cite{ma2025should, srinath2025assessing}.
Lastly, all users rated \model{} as fun and similar to existing tools, highlighting that user studies in our system can be engaging and ecologically valid \cite{brunswik1955representative}, i.e., mirroring downstream use.

% Overall, our pilot study reveals \model{} is effective and easy to use for agentic coding workflows.

% \input{data/site}

% One author with previous publications in website development then blindly judged three users' websites (one per task) compared to one-shot outputs from GPT-5.2, Sonnet 4.5, and Gemini-3 Pro using four metrics taken from a Game Jam competition: task adherence, creativity, style, and enjoyment.\footnote{https://itch.io/jam/gmtk-2024/results/top-marks}
% Despite using a weaker \mm{} (GPT-4.1), \model{}'s human–\ai{} teams made sites rated higher on subjective facets (style, creativity, enjoyment, Figure~\ref{fig:site}) beyond task completion---required for high-quality web design \cite{moshagen2010facets}.
% This preliminarily shows user–\ai{} collaboration in \model{} may surpass strong \mm{}s in open design tasks.

\begin{figure}[t!]
    % \vspace{-1ex}
    \centering
    {\includegraphics[width=\linewidth]{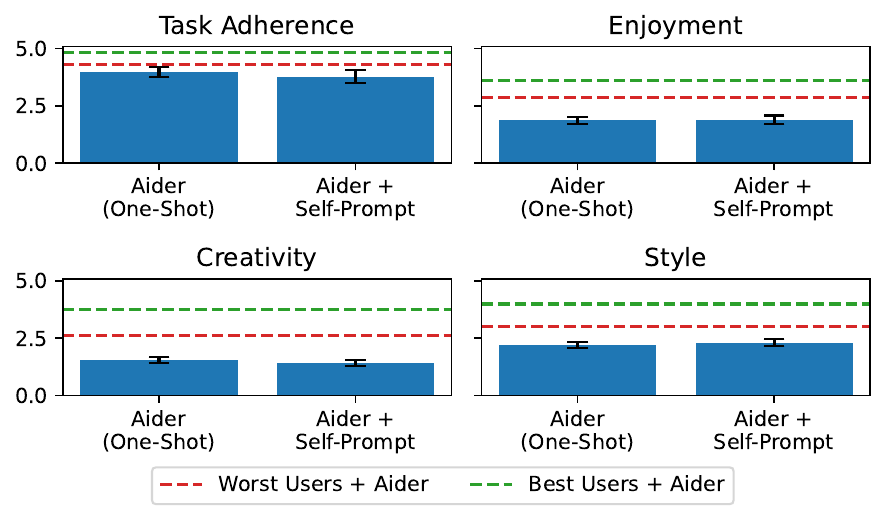}}
    \vspace{-4ex}
    \caption{Evaluation of game websites via \textsc{Aider} alone versus user+\textsc{Aider} teams. Our users beat the~agents in generating enjoyable, creative, and well-styled games.}
    % \vspace{-1ex}
    \label{fig:eval}
\end{figure}

\subsection{In-the-Wild Evaluation: Users Develop Better Games Compared to Agents Alone}
\label{subsection:wild}

To motivate future user-in-the-loop evaluations in \model{}, we test our system in a realistic setting.
$13$ junior programmers (CS undergrads)~use \model{} to make $22$ websites from our full list of games (\cref{subsection:research}), including platformer, puzzle, and reaction-time games.
Users prompt \textsc{Aider} $12.85 \pm 3.25$ on average per game, indicating high engagement.

We compare the user+\textsc{Aider} teams to two agent-only baselines: (1) prompting \textsc{Aider} once to create a website from the task description; and (2) taking this initial website and re-prompting \textsc{Aider} seven more times---the median number of prompts users ask---using its generated follow-up ideas (\cref{subsection:ai}).
Setting (2) controls for prompting effort.
\textsc{Aider} uses the same configuration as in the user study.

The first two authors score websites on 5-Likert scales via metrics from Game Jams: task adherence, enjoyment, creativity, and style.\footnote{\url{https://itch.io/jam/gmtk-2024/results/top-marks}}
Their Cohen's $\kappa$ is $0.23$, which indicates fair agreement \citep{cohen1960coefficient}.
This highlights the subjective nature of game evaluation, which is why Game Jams often use large-scale crowdsourcing. 
For our evaluation, we average the authors' scores to reduce variance.

Even the lowest-scoring user+\textsc{Aider} teams surpass both agent-only baselines on enjoyment, creativity, and style (Figure~\ref{fig:eval}).
The exception is task adherence, where most websites are perfect.
Thus, progress in coding agents requires metrics beyond just task completion and analyses of how users can improve outputs with agents \citep{shen2025completion}---gaps \model{} closes.
We plan to keep running this study across all of our $55$ game tasks and keep releasing user-agent interaction data as it accrues.

% \subsection{Case Study: Comprehension Questions} \label{subsection:question}

% We previously discussed \model{}'s potential for generating personalized post-submission questions (\cref{subsection:submission}), so we now empirically test it for code comprehension \cite{maalej2014comprehension}.
% Comprehension is often tested via manually-written code-specific questions \cite{wyrich202340}, so we now examine whether \mm{}s can generate these questions automatically.
% Using the websites produced in \cref{subsection:pilot}, we generate 27 multi-select comprehension questions with GPT-5.2. The questions ask participants to select 1) which features the website contains or does not contain; and 2) which JavaScript functions are present or absent.
% Our manual review confirms that \textit{all} generated questions are factually accurate.

% Users have a mean accuracy of $0.76$  ($0.875$ for website feature questions and $0.625$ for JavaScript ones), while random chance is $0.50$, so our preliminary results show users may not fully comprehend code in agentic workflows.
% The accuracy/feasibility of automatic question generation, and preliminary evidence that these questions can reveal interesting coding behaviors, show that \model{}'s personalized questions are a promising research direction.

\section{Related Work: Agentic Coding Tools} \label{section:related_work}

% \connor{worth a table for the lazy?}

Given \model{}'s focus on coding agents, we summarize the main ways researchers study such agents and how \model{} addresses their issues.
One~study type observes~users as they interact with existing, commercial systems like Github Copilot or Cursor \cite{barke2023grounded, chou2025building, shah2025students}.
While valuable for understanding current coding practices, researchers have no control of the agent and data collection requires annotation.
In contrast, \model{} is customizable and logs all data.

The second type forks IDEs like VSCode \cite{chi2025copilot}, including CodeGenie \cite{zhao2025codinggenie} for proactive \ai suggestions, Ivie for code explanations \cite{yan2024ivie}, and Trailblazer to answer questions in repositories \cite{yan2025answering}. 
However, these platforms require setup from all of the study participants, while researchers can host \model{} completely online to simplify onboarding.

% A third system type is completely private, recruiting participants within university courses \cite{vadaparty2024cs1, 10.1145/3708359.3712104} or online interview platforms \cite{shen2026ai}.
% Instead, \model{} is open-source and reproducible.

Lastly, systems closest to us also host online interfaces, such as
CodeHelp \cite{10.1145/3631802.3631830}, RealHumanEval \cite{mozannar2024realhumaneval, chen2025need}, CodingGenie \citep{zhao2025codinggenie}, EditTrail \citep{zhang2026editrail}, TiCoder \citep{lahiri2022interactive}, and Codellaborator \citep{pu2025assistance}.
However, such systems either do not release their code \citep{vadaparty2024cs1, 10.1145/3708359.3712104}, do not use modern agents (\cref{subsection:ai}), or study function completion---which researchers now value less (\cref{section:formative}).
\model{} addresses all of these issues.

% Two exceptions to this categorization are RealHumanEval \cite{mozannar2024realhumaneval} and CodeHelp \cite{10.1145/3631802.3631830}.
% However, their \ai support is limited to chat modes and auto-complete which cannot directly edit the user's code---how users increasingly interact with \ai for coding \cite{chen2025code}.
% \model{}'s \ai support is focused on agentic coding tools to align with modern \ai coding use.

\section{Conclusion: Ways to Jam in \model{}} \label{section:conclusion}

Through online studies with users and surveys with researchers, we reveal the value of \model{}:~a~system for \ai{}-assisted coding workflows.
We conclude with future research directions that were previously difficult to study, but more feasible with \model{}.

\vspace{0.75em}

\noindent \textbf{Long-Horizon Execution.} We increasingly want \ai systems to execute longer tasks \cite{task-completion-time-horizons-of-frontier-ai-models}, including programming.
However, as long-horizon execution requires intermediate feedback \cite{struber2025measuring, balepur2026dracula}, researchers could deploy \model{} to collect traces from expert programmers while they complete long tasks.
We can train models on how experts decompose tasks, which \ai{} changes are accepted, and which follow-up ideas are useful, aiding long-horizon execution.

\vspace{0.75em}

\noindent \textbf{Rewarding Code Quality.} The main reward for coding benchmarks is task completion \cite{chen2021codex, jimenez2024swebench}.
However, this alone may incentivize low-quality coding practices---like extensive fallbacks and hard-coded paths \cite{sabra2025assessing}---that preclude downstream use.
We are excited about the possibility of training reward models on programmer edits \cite{chakrabarty2025ai} or human judgments of code in \model{}'s evaluation pane (\cref{subsection:evaluation}) to improve \ai code quality. 

\vspace{0.75em}

\noindent \textbf{Monitorability.} A safety risk of coding agents is monitorability \cite{korbak2025chain}; programmers may struggle to identify severe \ai bugs, especially if they have not learned the skills to do so \cite{shen2026ai}.
\model{} could serve as a platform to train these skills: teaching programmers how to~quickly and accurately review \ai-generated code \citep{10.1145/3660764}, designing scaffolds to make monitoring simpler \citep{guan2026monitoring}, and building tools to automatically detect failures from system execution traces~\citep{balepur-etal-2026-test, balepur-etal-2026-benchmarker}. 
\vspace{0.75em}

\noindent \textbf{Agent Design.} The best coding agent designs are opaque \cite{wang2025humansmissing} and without a platform to evaluate them, researchers must follow decisions from commercial tools.
\model{} changes this by letting researchers test custom agents, answering: which \mm{} best improves productivity \cite{chi2025copilot}, the value of planning and clarification pre-execution \cite{balepur2025good}, and methods for handling long-context inputs \cite{zhang2025recursive}.

\section*{Acknowledgments}

We thank the \abr{clip} lab at the University of Maryland for their support.
In particular, we thank Yapei Chang, Zongxia Li, Dang Nguyen, Feng Gu, Yu Hou, Paiheng Xu, and Maharshi Gor for demoing our system.
This material is based upon work supported by the National Science Foundation under \abr{iis}-2339746 (Rudinger), \abr{iis}-2403436 (Boyd-Graber), and \abr{dge}-2236417 (Balepur).
Boyd-Graber's research is also supported by a gift from Adobe Corporation.
Any opinions, findings, and conclusions or recommendations expressed in this material are those of the author(s) and do not necessarily reflect the views of sponsors.
 
% \section{Limitations} \label{section:limitations}

% The ui is perfect there are none

% \section{Ethical Considerations}

% None

% \section*{Acknowledgments}

% We would like to thank the \abr{clip} lab at the University of Maryland and our external collaborators for their help.
% This material is based upon work supported by the National Science Foundation under \abr{iis}-2339746 (Rudinger) \abr{iis}-2403436 (Boyd-Graber), and \abr{dge}-2236417 (Balepur).
% Any opinions, findings, and conclusions or recommendations expressed in this material are those of the author(s) and do not necessarily reflect the views of the National Science Foundation.
% Access to Cohere models (Command-R, Command-R Plus)~was made possible via a Cohere for AI Research Grant.

\bibliography{custom}
\bibliographystyle{acl_natbib}

\clearpage

\appendix
\section{Appendix} \label{section:appendix}

\subsection{Case Study: Comprehension Questions} \label{subsection:question}

We previously discussed \model{}'s potential for generating personalized post-submission questions (\cref{subsection:submission}), so we now empirically test it for code comprehension \cite{maalej2014comprehension}.
Comprehension is often tested via manually-written code-specific questions \cite{wyrich202340}, so we now examine whether \mm{}s can generate these questions automatically.
Using the websites produced in \cref{subsection:pilot}, we generate 27 multi-select comprehension questions with GPT-5.2. The questions ask participants to select 1) which features the website contains or does not contain; and 2) which JavaScript functions are present or absent.
Our manual review confirms that \textit{all} generated questions are factually accurate.

Users have a mean accuracy of $0.76$  ($0.875$ for website feature questions and $0.625$ for JavaScript ones), while random chance is $0.50$, so our preliminary results show users may not fully comprehend code in agentic workflows.
The accuracy/feasibility of automatic question generation, and preliminary evidence that these questions can reveal interesting coding behaviors, show that \model{}'s personalized questions are a promising research direction.

\subsection{Prompts}

In Prompt~\ref{prompt:summary_and_idea} and Prompt~\ref{prompt:question_generation}, we provide our custom prompts for generating summaries and follow-up ideas from agent executions, and personalizing questions based on a user's codebase, respectively.

\hypersetup{
    linkcolor=white,
    citecolor=white,
    urlcolor=white
}

\lstset{
  literate={<}{{<}}1
           {>}{{>}}1
}

\begin{prompt}[title={Prompt \thetcbcounter: Summary and Idea Generation Prompt (\cref{subsubsection:ai_execution})}, label=prompt:summary_and_idea]
You are an expert at summarizing actions that an AI assistant took after being prompted by a user and providing useful suggestions for the user to improve their code.\\

This is what the user asked the assistant to do:\\
<query>\\
\texttt{[insert user query]}\\
</query>\\

These are the final versions of files after edits (only changed files included):\\
<final\_files>\\
\texttt{[insert final files]}\\
</final\_files>\\

These are the changes that the assistant made to the code (with optional SEARCH/REPLACE edit blocks when available):\\
<changes>\\
\texttt{[insert edits]}\\
</changes>\\

Using this information your job is to generate:\\
1. A summary of the changes that the assistant made to the code.\\
2. A list of ideas for the user to improve their code.\\

<summary instructions>\\
- The summary should be written in first person as if you were the one who made edits to the code. Use "I" as appropriate.\\
- You must discuss which files were edited and the specific changes to each file.\\
- Be subtle in how the changes address the user's request; do not quote the user's request.\\
- Be concise. The summary should be a maximum of two sentences.
</summary instructions>\\

<idea instructions>\\
- Generate 3 ideas with their corresponding probabilities, sampled from the full distribution.\\
- Each idea should improve the code or the task: e.g. task fulfillment, correctness, style, readability, edge cases, or user experience, depending on what fits the project (web UI, Python script, etc.).\\
- Only suggest ideas that are feasible given the file types and stack (e.g. for web: HTML/CSS/JS; for Python: standard library, tests, clarity). Do not suggest custom assets, external services, or out-of-scope changes.\\
- Frame each idea as a follow-up action the user could ask for, i.e. a short command starting with a verb.\\
- Be concise. Each idea should be no more than 10 words.\\
</idea instructions>\\

<format instructions>\\
Generate your output as a json with two keys: 1) "summary" with a string value of the summary; 2) "ideas" with a list of strings value of the ideas; and 3) "probabilities" with a list of floats value of the probabilities of each idea based on your full distribution.\\
{{\\
    "summary": "insert summary",\\
    "ideas": ["insert idea 1", "insert idea 2", "insert idea 3"],\\
    "probabilities": [float probability 1, float probability 2, float probability 3],\\
}}\\
Do not generate anything else\\
</format instructions>
\end{prompt}

\begin{prompt}[title={Prompt \thetcbcounter: Chat Mode Prompt (\cref{subsubsection:ai_execution})}, label=prompt:chat]
You are VibeJam Chat Mode: a helpful coding assistant.\\

You will receive a user's question plus their current project files. Your job is to help the user by:\\
- answering their question with respect to their code base\\
- providing brief, relevant code snippets \\

Hard rules (Chat Mode):\\
- Do NOT apply edits, do NOT claim you changed files, do NOT output patch/search-replace blocks.\\
- Do NOT output "SEARCH/REPLACE" or Aider-style replacement snippets.\\
- Only respond in natural language and optional fenced code blocks.\\
- If you include code, use fenced blocks with an appropriate language tag like ```html, ```css, ```javascript.\\

Encouragement:\\
- If the users' query indicates that they actually want to make edits automatically, suggest switching to Agent mode.\\

User request:\\
<user\_request>\\
\texttt{[insert user prompt]}
</user\_request>

Current project file context (trimmed):
<files\_context>
\texttt{[insert files]}
</files\_context>
\end{prompt}

\begin{prompt}[title={Prompt \thetcbcounter: Plan Mode Prompt (\cref{subsubsection:ai_execution})}, label=prompt:plan]
You are a planning assistant that writes and revises implementation plans for a request to implement a feature.\\

User request:\\
<user\_request>\\
\texttt{[insert user query]}\\
</user\_request>\\

Here is the previous plan you may need to revise:\\
<previous\_plan>\\
\texttt{[insert prior plan, blank if nonexistent]}\\
</previous\_plan>\\

Current project file context (trimmed):\\
<files\_context>\\
\texttt{[insert files]}\\
</files\_context>\\

Instructions:\\
- Return a full revised plan document in markdown.\\
- If the previous plan is blank, create a complete first plan.\\
- If the previous plan exists, integrate the new user request by revising it.\\
- Keep the output implementation-focused and actionable.\\
- Do not wrap the result in code fences.\\
- Tailor the length of the plan based on the complexity of the request. Simpler requests can have shorter plans; complex requests can have longer plans\\
- Eventually, this plan will be fed into a coding agent that will be tasked to execute all of the requests in the plan. So make sure that the plan is clear such that someone could follow it and then execute it. For this same reason, you only need to focus on implementation steps, not testing, etc.\\

The format of the plan should be as follows:\\
<format>\\
\#\# Plan: [summarized tl;dr of feature request]\\

\#\#\#\# Objective: [insert objective]\\

\#\#\#\# Implementation Steps:\\
1. numbered list of steps\\
...\\
n.\\

\#\#\#\# Constraints and Notes:\\
- bullet point list of steps\\
...\\
</format>\\
If you believe that any of these sections are not necessary, do not include them
\end{prompt}

\begin{prompt}[title={Prompt \thetcbcounter: Plan Summarization Prompt (\cref{subsubsection:ai_execution})}, label=prompt:plan_summary]
You summarize how a markdown plan changed after a user request.\\

User request:\\
<user\_request>\\
\texttt{[insert user query]}\\
</user\_request>\\

Previous plan:\\
<previous\_plan>\\
\texttt{[insert prior plan, blank if nonexistent]}\\
</previous\_plan>\\

Updated plan:\\
<updated\_plan>\\
\texttt{[insert current plan]}\\
</updated\_plan>\\

Rules:\\
- Summarize only what was added, removed, or changed in the updated plan.\\
- Do not propose follow-up ideas or next-step suggestions.\\
- Keep it concise (max 3 short sentences).\\
- Return plain text only.\\
- Use first person to indicate that you wrote the plan.\\

At the end of the summary, briefly hint that the user can either keep making adjustments to the plan or can hit "Build" to execute the plan\\
\end{prompt}

\begin{prompt}[title={Prompt \thetcbcounter: Question Generation Prompt (\cref{subsection:question})}, label=prompt:question_generation]
<task>\\
Given the user's HTML, CSS, and JavaScript code, generate five features that exist and five that do not but plausibly could.\\
</task>\\
<html>\texttt{[insert HTML]}</html>\\
<css>\texttt{[insert CSS]}</css>\\
<javascript>\texttt{[insert JS]}</javascript>\\
<format>\\
{{ "real\_features": ["...", ...], "fake\_features": ["...", ...] }}\\
</format>
\end{prompt}

\end{document}